\documentclass[11pt,a4paper]{article}
\usepackage[hyperref]{acl2019}
\usepackage{times}
\usepackage{latexsym}
\usepackage{amsmath}
\usepackage{amsfonts}
\usepackage{graphicx}
\usepackage{booktabs}
\usepackage{url}
\usepackage{graphicx}

\aclfinalcopy % Uncomment this line for the final submission
\title{Charge-Based Prison Term Prediction with Deep Gating Network}
\author{Huajie Chen$^1$\thanks{~~Both authors contributed equally.}~~~~Deng Cai$^{2*}$~~~Wei Dai$^1$~~~Zehui Dai$^1$~~~Yadong Ding$^1$\\$^1$NLP Group, Gridsum, Beijing, China\\{\tt \{chenhuajie,daiwei,daizehui,dingyadong\}@gridsum.com}\\$^2$The Chinese University of Hong Kong \\ {\tt thisisjcykcd@gmail.com}}

\date{}

\begin{document}
\maketitle
\begin{abstract}
  Judgment prediction for legal cases has attracted much research efforts for its practice use, of which the ultimate goal is prison term prediction. While existing work merely predicts the total prison term, in reality a defendant is often charged with multiple crimes. In this paper, we argue that charge-based prison term prediction (CPTP) not only better fits realistic needs, but also makes the total prison term prediction more accurate and interpretable. We collect the first large-scale structured data for CPTP and evaluate several competitive baselines. Based on the observation that fine-grained feature selection is the key to achieving good performance, we propose the Deep Gating Network (DGN) for charge-specific feature selection and aggregation. Experiments show that DGN achieves the state-of-the-art performance.
\end{abstract}
\section{Introduction}
\begin{figure}
  \centering
  \includegraphics[width=7.5 cm]{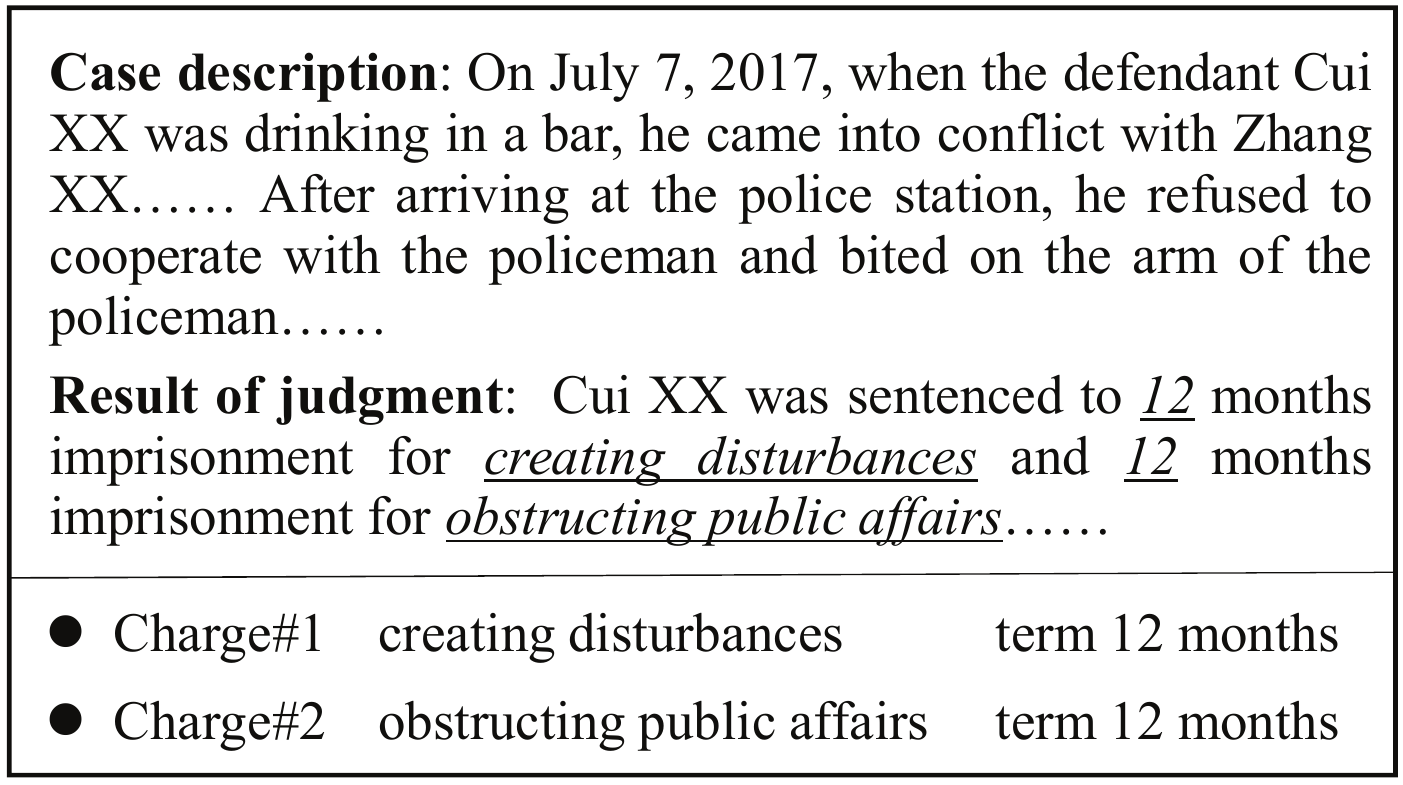}\\
  \caption{An example of judgment prediction.}\label{example}
\end{figure}

Judgment prediction \cite{kort1957predicting, ulmer1963quantitative,segal1984predicting,liu2004case,liu2006exploring} aims at automatically predicting the judgment result given a textual description of a legal case (An example is given in Figure \ref{example}). Recently, there has been a resurgent interest in this task due to the availability of more data and new machine learning techniques \cite{D17-1289, D18-1390, hu2018few}.

Judgment prediction can be decomposed into several sub-tasks: (a) relevant law article extraction~\cite{liu2006exploring, liu2005classifying, liu2015predicting}, (b) charge prediction \cite{liu2006exploring, D17-1289, hu2018few}, (c) and prison term prediction~\cite{zhong2018overview}. The dependencies among them have also been studied by \newcite{D18-1390}. While effective methods exist for sub-task (a) and (b), (e.g In CAIL2018 competition \cite{zhong2018overview}, both the charge prediction and the article prediction have attained $F_{\text{micro}}$ over $95\%$), the prison term prediction remains the performance bottleneck.

In this paper, we improve the accuracy of prison term prediction by decomposing it into a set of \textit{charge-based} prison term predictions (CPTPs). In this way, more subtle and sophisticated interactions between textual description and a specific charge can be captured, resulting in more precise term predictions for individual charges. Meanwhile, CPTPs also shed light on the prediction of the total prison term.

On the other hand, CPTP also poses challenges due to the following reasons: The case description can be very lengthy and not all parts are relevant to a specific charge. The charge-related descriptions are often presented in an interleaving way, making it difficult to associate a specific charge with its corresponding information.

To address the above problems, we propose the Deep Gating Network (DGN) for gradually filtering and aggregating charge-specified information at different levels of granularity. Specifically, we stack multiple blocks of an LSTM layer and a charge-specific gating layer for generating a focused charge-based representation of the case description. Finally, the whole document representation is obtained by a convolutional neural network.

To conduct the experiments, we construct a new dataset, which contains more than 200, 000 criminal cases.\footnote{The dataset can be found at \url{https://github.com/huajiechen/CPTP}} To show the effectiveness of the proposed approach, we compare it with several strong baselines adapted from aspect-based sentiment classification \cite{D16-1058, D16-1021, D17-1047, P18-1087}. Experiments show that our method achieves significantly better results than all of them. In addition, when we leverage the results of charge-based term predictions for the total prison term prediction, it also surpasses several strong baselines that are directly aimed at the total term prison prediction.

In summary, our contributions are as follows:
\begin{itemize}
\item We formally define the task of charge-based prison term prediction and collect the first dataset for it.
\item We propose the Deep Gating Network (DGN). Experiments show our method achieves the state-of-the-art performance.
\item We show that the accuracy of the total term prediction is also improved by a simple heuristic integration of individual charge-based term predictions.
\end{itemize}

\section{Problem Definition \& Dataset Construction}
We formally define the task of charge-based prison term prediction as follows. The input are a case description $\mathbf{x}=\{x_1, x_2, \cdots, x_n\}$ and a set of corresponding charges $\mathbf{c}=\{c_1, c_2,\cdots, c_k\}$, where $n$ and $k$ are the length of case description and the number of charges respectively. The goal is to predict the prison terms $\mathbf{y}=\{y_1, y_2,\cdots, y_k\}$, where $y_j$ is the prison term corresponding to charge $c_j$.

To the best of our knowledge, there is no existing structured dataset for the above task. We thus collect and construct a dataset based on the published records from the Supreme People's Court of China,\footnote{ \url{http://wenshu.court.gov.cn/}} where each criminal case document includes the accusation by the procuratorate, the court view, and the result of judgment. Following \newcite{xiao2018cail2018}, we take the accusation by the procuratorate as the input textual description. The charges and the corresponding prison terms are extracted from the result of judgment using regular expressions like ``sentence to \underline{~~~~}  months imprisonment for \underline{~~~~}''. We build 238,749 well-structured cases in total (An example is given in Fig \ref{example}). The collected cases are further split into the training set, the validation set, and the test set. The statistics of the dataset are detailed in Table \ref{stat}. The range of possible prison terms is [1, 240] (in months). The dataset has a broad coverage of common charges, 157 different types of charges are involved.
\begin{table}[t!]
\begin{center}
\begin{tabular}{lrrr}
\toprule
      & \multicolumn{1}{c}{\#single} & \multicolumn{1}{c}{\#multiple} & \multicolumn{1}{c}{total} \\ \midrule
Train & 147, 580                       & 42, 420                          & 190, 000                    \\
Valid & 19, 350                        & 5, 602                           & 24, 952                     \\
Test  & 18, 539                        & 5, 258                           & 23, 797    \\  \bottomrule
\end{tabular}
\end{center}
\caption{\label{font-table} The statistics of the proposed dataset. \#single/\#multiple means single/multiple charge(s) cases.}
\label{stat}
\end{table}
\section{Our Approach}
\begin{figure}
  \centering
  \includegraphics[width=6.5cm]{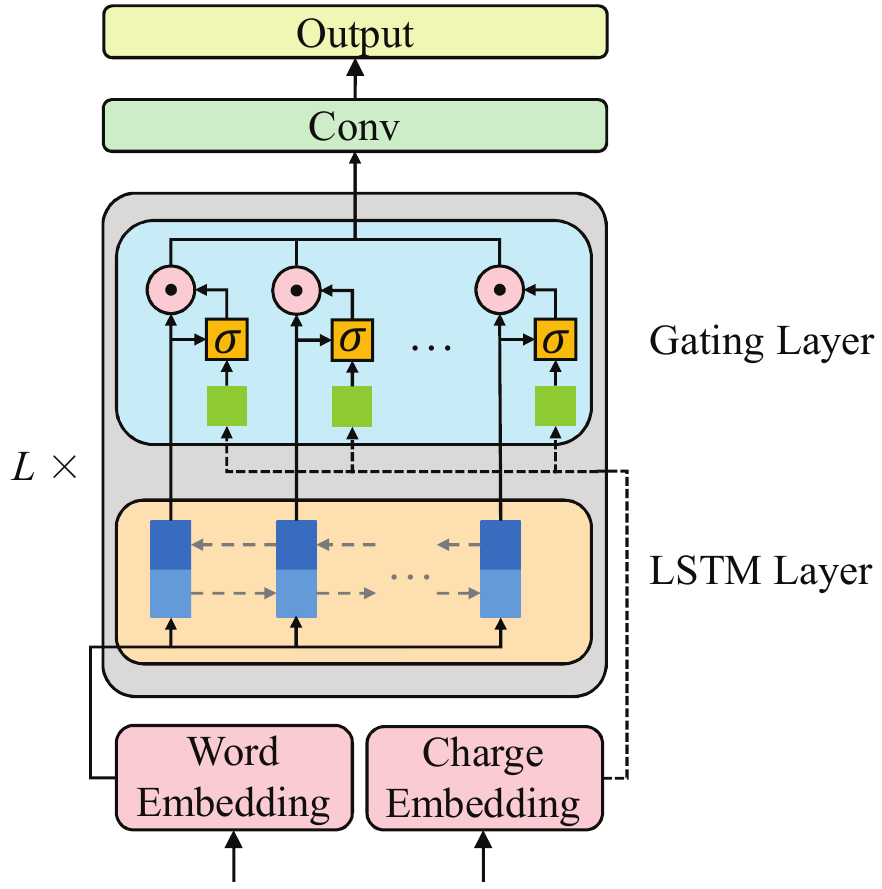}\\
  \caption{The architecture of DGN}\label{figure architecture}
\end{figure}
Figure \ref{figure architecture} gives an overview of our model, which consists of two components: (1) Deep Gating Network (DGN) for charge-based feature filtering and aggregation. (2) Convolutional Neural Network (CNN) for the whole document representation learning.
\subsection{Deep Gating Network}
At the bottom layer of DGN, each word $x_i$ is mapped into a low dimensional vector $h^{(0)}_i$ according to a word embedding table.

DGN then starts to construct charge-specific representations gradually. $L$ identical blocks are hierarchically stacked. The $l$-th block takes the output of the $(l-1)$-th layer $h_i^{l-1}$ as input. Each block transforms its input semantic vectors into more sophisticated and focused representations based on gated feature selection and combination.

Specifically, each gating block consists of a bi-LSTM layer for context aggregation and a gating layer for charge-specific feature filtering.
\begin{gather}
\tilde{h}^{(l)}_i=[\overrightarrow{\text{LSTM}}(h^{(l-1)}_i);  \overleftarrow{\text{LSTM}}(h^{(l-1)}_i)] \notag \\
h^{(l)}_k=g^{(l)}_i\odot \tilde{h}^{(l)}_i \notag
\end{gather}
where $g^{(l)}_i$ is the gate for $i$-th vector of the $l$-th gating block and $\odot$ denotes element-wise multiplication. $g^{(l)}_i$ is computed as:
\begin{equation}
g^{(l)}_i=\text{sigmoid}{(W^{(l)}[\tilde{h}^{(l)}_i;c_j]+b^{(l)})} \notag
\end{equation}
where $c_j$ is target charge embedding. The gating layers can select the charge-specific features according to the target charge embedding.
\subsection{Convolutional Neural Network}
Convolutional Neural Network (CNN) has been effective in modeling sequential data \cite{D14-1181, hu2014convolutional, pang2016text}. It uses convolution operations (with multiple groups of filters) for $n$-gram feature extraction. The sequence-level representation is then obtained through max-pooling, where the most salient $n$-gram features are detected and selected.

In this work, we use a CNN with filter width in $[1, 2, 3, 4, 5]$. The number of filters for each width is 256. We concatenate the outputs of different filters for the final document representation $z$.
\subsection{Output and Training}
The charge-specific document representation $z$ is passed to a fully connected layer with ReLU activation for the final prediction.
\begin{equation}
\hat {y}_{j} = \text{ReLU}(W_o z + b_o)
\label{output}
\end{equation}
where $W_o$ and $b_o$ are trainable parameters.

Since the Mean Squared Error (MSE) loss cannot reflect the relative deviation ratio between the prediction and the ground-truth, we take the logarithm before estimating their difference.
\begin{equation}
\delta_j = |\log(y_j+1)-\log(\hat {y}_j+1)|
\label{delta}
\end{equation}

To alleviate the impact of outliers and stabilize the training,
we propose to use Huber Loss \cite{Huber1964Robust}, $a$ is set to 1 in experiments:
\begin{gather}
l(x, y_j)=\begin{cases}
0.5 \times \delta_j^2,  &\text{if $\delta_j < a$}   \\
a(\delta_j-0.5a),  &\text{otherwise} \\
\end{cases} \notag \\
\mathcal L(x, y) = \sum_{j=1}^{k}l(x, y_j) \notag
\end{gather}
\paragraph{Total Term Prediction}
Although our model is trained to predict the prison term for specific charge, it can be readily adapted to predict the total term by a simple heuristic integration of individual charge-based prison term predictions. There are certain regulations for combined punishment of crimes in Chinese legislation. For simplicity, we take the average of the maximum and summation of individual charge-specific term predictions. The total term prediction is also capped at 240 months.
\begin{equation}
\begin{split}
\label{total term}
    \hat{y}_{total}=\min(240,\frac{\max_j(\hat{y}_j) + \sum_j(\hat{y}_j)}{2}) \notag
    %\hat{y}_{total}=\frac{\max_j(\hat{y}_j) + \sum_j(\hat{y}_j)}{2} \notag
\end{split}
\end{equation}

\section{Experiments}
\subsection{Evaluation Metrics}
For evaluation, we adopt the official score function (S metric) of the CAIL2018 Competition \cite{zhong2018overview}. The score function measures the log different $\delta$ between prediction value $\hat{y}$ and gold value $y$ as in Eq \ref{delta}. The final score $s(\delta)$ is a piece-wise function that increases monotonically with the value of $\delta$. For more details about the S metric, we refer interested readers to \cite{zhong2018overview}.
We also report the exact match (EM) rate and error-tolerant accuracy Acc@$p$, where $p$ is the maximum acceptable error rate. Formally, a prediction is considered ``correct" if and only if its value is in the range $[{y}(1-p),  {y}(1+p)]$.
\subsection{Compared Methods}
The task of charged-based prison term prediction is similar in spirit to aspect-based sentiment classification \cite{pang2008opinion}, where multiple classification decisions are made given one text description and different target entities. This suggests that other neural architectures proposed for aspect-based sentiment classification may also be suitable for our task. The adaption from classification to regression can be easily accomplished by replacing the original final layer with that of Eq \ref{output}. Specifically, we adapted the following models:
\begin{itemize}
\item ATAE-LSTM \cite{D16-1058}: it concatenates aspect embedding and the output of LSTM, and uses self-attention to obtain aspect-based representation.
\item MemNet \cite{D16-1021}: it uses multi-hop attention over the word embeddings for a sentence, where aspect embedding is regarded as the initial key.
\item RAM \cite{D17-1047}: it also uses multi-hop attention for aspect-specific representation learning, while the attention at different time steps are aggregated by recurrent neural network.
\item TNet \cite{P18-1087}: it has a similar architecture to DGN. The major difference is that it employs a Transformation Network for mixing the information in aspect embedding and token representations rather than the explicit gates in our model.
\end{itemize}
The aspect embedding in above models is replaced by charge embedding in our experiments. In addition, we also compare with the popular models for total term prediction \cite{zhong2018overview, D18-1390}:
\begin{itemize}
\item CNN \cite{D14-1181}: the case description is encoded by a CNN with multiple filter widths, followed by max-pooling.
\item RNN \cite{hochreiter1997long}: bi-LSTM are used for case description encoding, where the final states are regarded as the document representation.
\item RCNN \cite{lai2015recurrent}: we stack a CNN on the top of LSTM states for final representation.
\end{itemize}
\subsection{Main Results}
\begin{table}[t!]
\begin{center}
\scalebox{0.9}{
\begin{tabular}{lcccc}
\toprule
Model & S & EM & Acc@0.1 & Acc@0.2 \\ \midrule
ATE-LSTM & 66.49 & 7.72 & 16.12 & 33.89 \\
MemNet & 70.23 & 7.52 & 18.54 & 36.75 \\
RAM & 70.32 & 7.97 & 18.87 & 37.38 \\
TNet & 73.94 & 8.06 & 19.55 & 39.89 \\ \hline
DGN & \textbf{76.48} & \textbf{8.92} & \textbf{20.66} & \textbf{42.61} \\ \bottomrule
\end{tabular}}
\end{center}
\caption{\label{Main result} Results on charge-based prison term prediction(\%). }
\end{table}
\begin{table}[t!]
\begin{center}
\scalebox{0.9}{
\begin{tabular}{lcccc}
\toprule
Model & S & EM & Acc@0.1 & Acc@0.2 \\ \midrule
CNN  & 67.24 & 8.41 & 16.96 & 35.58 \\
RNN  & 67.27 & 8.04 & 16.79 & 35.11 \\
RCNN & 69.56 & 8.54 & 17.57 & 35.75 \\ \hline
DGN & \textbf{75.74} & \textbf{8.64} & \textbf{19.32} & \textbf{40.43} \\ \bottomrule
\end{tabular}}
\end{center}
\caption{\label{Main result2} Results on total term prediction(\%). }
\end{table}
The results of charge-based prison term prediction are shown in Table \ref{Main result}. The proposed DGN achieves the best results on all four metrics. In addition, the margins between our model and others are remarkably wide. It can be observed that aspect-based sentiment models only give moderate performance, which we attribute to that the case description is so long that more rigorous feature selection, such as the treatment of DGN, is needed. Our model selects and aggregates features in a explicit way which is more efficient and effective in dealing with charge-specific descriptions often spread out across lengthy case documents in CPTP.

Table \ref{Main result2} presents the results of the total term prediction. Although our method is not directly trained to make the final prediction, the performance of our model surpasses all baselines, which confirms that the breakdown charge-based analysis can indeed help the total prison term prediction.
\subsection{Depth of DGN}
\begin{figure}
  \centering
  \includegraphics[width=8 cm]{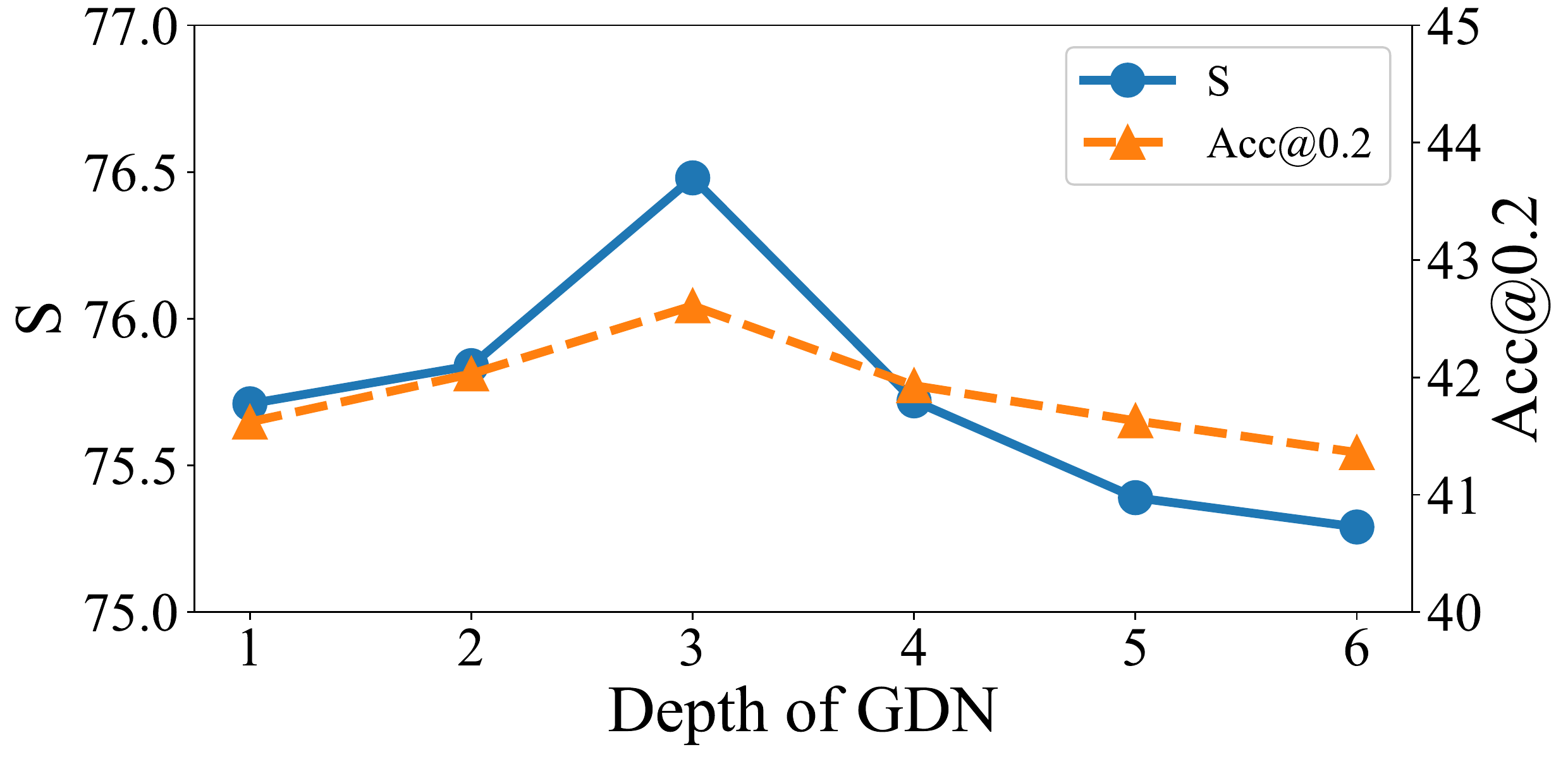}\\
  \caption{Performance with different depths of DGN.}\label{layer}
\end{figure}
To study the impact of the number of DGN blocks, we test our model with various depths and show the results in Fig \ref{layer}.\footnote{For simplicity, we only show S score and Acc@0.2.} As shown, the performance improves as the depth of DGN increases until it reaches 3 when the performance begins to drop likely due to overfitting.
\subsection{Effects of Log Huber Loss}
 \begin{figure}
   \centering
   \includegraphics[width=7.5 cm]{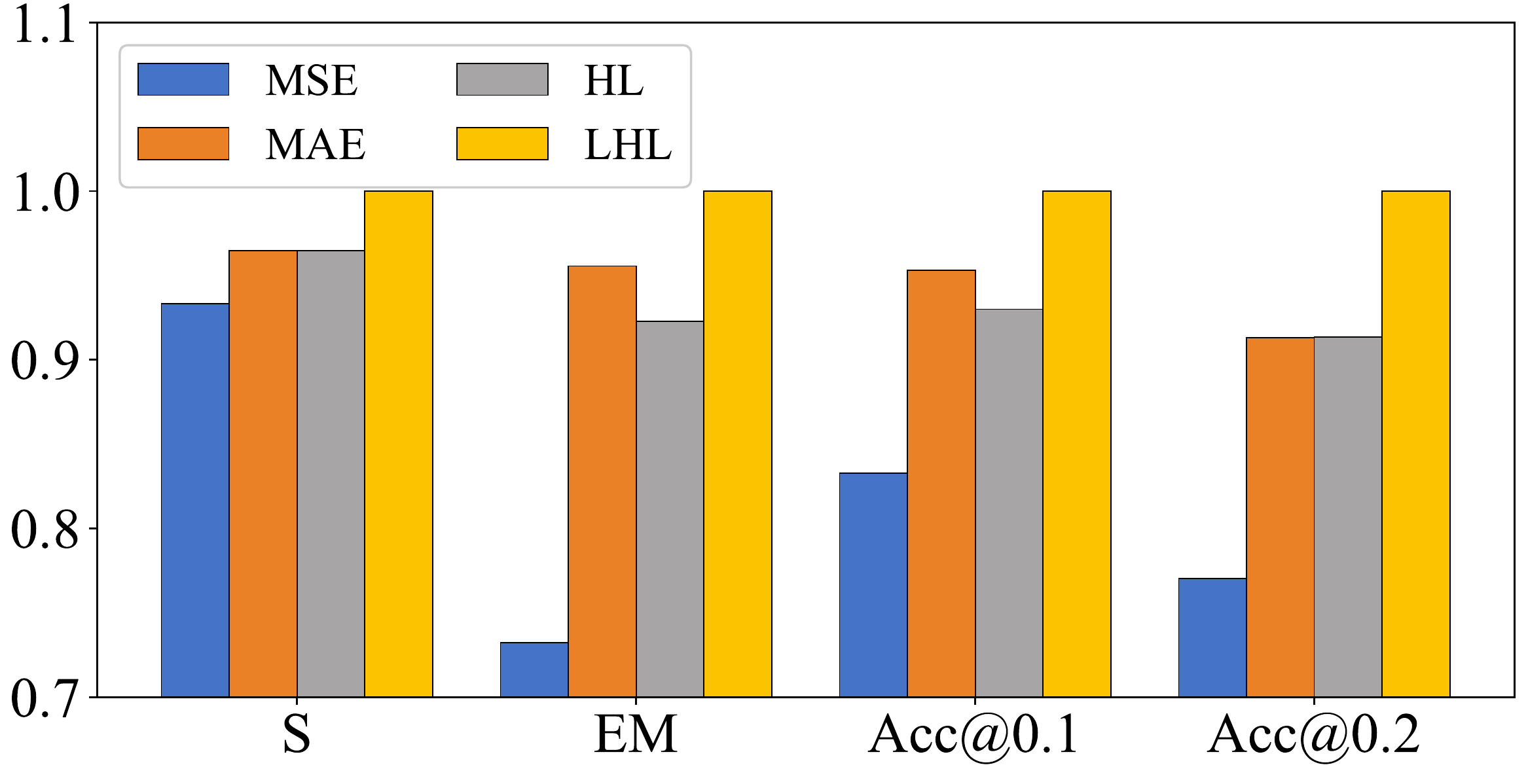}\\
   \caption{Compare of different loss functions. The result of LHL is regarded as unit 1.}\label{loss_compare}
 \end{figure}
We compare Log Huber Loss (LHL) with Mean Square Error (MSE), Mean Absolute Error (MAE) and Huber Loss (HL). We also try Log Cosh Loss (LCL), but it does not converge.
As shown in Figure \ref{loss_compare}, Log Huber Loss performs best in all metrics for all models. The improvement is most significant in S metric. It also suggests that making the loss function consistent with the evaluation metric is beneficial.
\subsection{Error Analysis}
So far, our model has the best results on prison term predictions. In this section, we aim to conduct an in-depth analysis and answer the following questions: (1) In which cases, our model fails to deliver accurate predictions? (2) What are the prospects for further improvement?  After carefully analyzing 100 examples, we roughly classify them into the following categories.
\paragraph{Lengthy Description} Some cases are extremely complicated, especially for cases with gangs. These descriptions are often lengthy and involve multiple criminal suspects.
\paragraph{Incomplete Information} In some cases, the input case description does not contain sufficient information for precise prediction. Note we only take the accusation by the procuratorate as input, which is incomplete compared to the whole materials relevant to a case. For example, if a defendant is recidivism within a shorter period, he/she shall be given a heavier punishment.
\paragraph{Rare Cases} Some special circumstances will influence the prison term, yet rarely happen in the training set. For example, if a defendant cause injuries to others due to excessive defense, he/she shall be given a lighter punishment. This knowledge is easily understandable by humans, bu hard to be learned by machine learning models.
\section{Ethical Discussions}
Although the research on prison term prediction has considerable potential to improve efficiency and fairness in criminal justice, there are certain ethical concerns worth discussions.

First, does the training data provide unbiased examples and sufficient? For example, some may worry about that the model may treat people differently based on race, social class, age and so on \cite{Tonry2014Legal}. Discrimination in the past may be learned in models. Also, with the development of our society, new forms of crimes will appear. A model trained on historical data may fail in these new cases.

Second, is the learned system robust enough? Some subtle details may significantly affect the result of judgment. For example, the amount of theft and the number of drugs, these numerical values are often not uniform in different case descriptions, causing it hard to learn by neural models. Some infrequent words, such as named entities, may also cause undesirable interference.

The mistake of legal judgment is serious, it is about people losing years of their lives in prison, or dangerous criminals being released to reoffend. We should pay attention to how to avoid judges' over-dependence on the system. It is necessary to consider its application scenarios. In practice, we recommend deploying our system in the ``Review Phase'', where other judges check the judgment result by a presiding judge. Our system can serve as one anonymous checker.

In summary, the judgment prediction is an emerging technology at its exploratory stage. We should be aware of the risks and prevent any inappropriate use of the technology.
\section{Conclusion}
In this paper, we formally presented the task of charge-based prison term prediction. We introduced the first large-scale dataset for this task. To tackle the problem of the noisy and entangled description of legal cases, we proposed the deep gating network for charge-specific information filter. Experiments show that our model significantly improves the accuracy of charge-based prison term prediction, as well as the total term prediction. Finally, we discussed some ethical problems of the proposed techniques that are worth cautious thinking.
\bibliography{acl2019}
\bibliographystyle{acl_natbib}
\end{document}